%% file: main_arxiv.tex
\pdfoutput=1

\documentclass[11pt]{article}

\usepackage{acl}

\usepackage{times}
\usepackage{latexsym}
\usepackage{graphicx}
\graphicspath{ {./images/} }
\usepackage{amsmath}
\usepackage{amssymb}
\usepackage{makecell}
\usepackage{pifont}
\newcommand{\cmark}{\ding{51}}%
\newcommand{\xmark}{\ding{55}}%
\usepackage[T1]{fontenc}

\usepackage[utf8]{inputenc}
\usepackage{booktabs,colortbl,multirow}

\usepackage[normalem]{ulem}
\usepackage{enumitem}
\usepackage{microtype}
\usepackage[disable]{todonotes}

\newcommand{\cneeded}[2]{\textcolor{orange}{[citation-needed]}}

\title{Large Language Models with Controllable Working Memory}

\author{{\bf Daliang Li$^{\spadesuit}$, Ankit Singh Rawat$^{\spadesuit}$, Manzil Zaheer$^{\heartsuit}$,Xin Wang$^{\spadesuit}$} \\ \bf{Michal Lukasik$^{\spadesuit}$, Andreas Veit$^{\spadesuit}$, Felix Yu$^{\spadesuit}$, Sanjiv Kumar$^{\spadesuit}$} \\
\normalsize{ $^\spadesuit$Google Research $^\heartsuit$Deepmind} \\ 
{\normalsize{\tt\{daliangli, ankitsrawat, manzilzaheer, wanxin\}@google.com}} \\
{\normalsize{\tt\{mlukasik, aveit, felixyu, sanjivk\}@google.com}} 
}

\begin{document}
\maketitle
\input{abstract}

\input{intro}

\input{contributions}

\input{related.tex}

\input{methods.tex}

\input{results}

\input{conclusion}

\input{future_work}

\bibliography{references} 

\appendix

\input{appendix}

\end{document}

%% file: abstract.tex
\begin{abstract}
Large language models (LLMs) have led to a series of breakthroughs in natural language processing (NLP), owing to their excellent understanding and generation abilities. Remarkably, what further sets these models apart is the massive amounts of world knowledge they {internalize} during pretraining. While many downstream applications provide the model with an informational context to aid its performance on the underlying task, how the model’s world knowledge interacts with the factual information presented in the context remains {under explored}. As a desirable behavior, an LLM should give precedence to the context whenever it contains task-relevant information that conflicts with the model's memorized knowledge. 
This enables model predictions to be grounded in the context, which can then be used to update or correct specific model predictions without frequent retraining.
By contrast, when the context is irrelevant to the task, the model should ignore it and fall back on its internal knowledge. In this paper, we undertake a first joint study of the aforementioned two properties, namely \textit{controllability} and \textit{robustness}, in the context of LLMs. We demonstrate that state-of-the-art T5 and PaLM (both pretrained and finetuned) could exhibit poor controllability and robustness, which do not scale with increasing model size. As a solution, we propose a novel method – \textbf{k}nowledge \textbf{a}ware \textbf{f}ine\textbf{t}uning (KAFT) – to strengthen both controllability and robustness by incorporating counterfactual and irrelevant contexts to standard supervised datasets. Our comprehensive evaluation showcases the utility of KAFT across model architectures and sizes.
\end{abstract}

%% file: intro.tex
\section{Introduction}
\label{sec:intro}
Large language models (LLMs) pretrained on large scale datasets have shown promising results across natural language tasks \citep{transformer, bert, t5, gpt3, gopher, palm, megatron}. However, as models scale ever larger, they become more expensive to train, making it unrealistic to frequently change model parameters. In real world applications, it is often necessary to adjust the model's behavior. This dilemma is especially sharp in the case of factual (world) knowledge that plays important role in realizing impressive performance of LLMs. It is well known that LLMs memorize large amounts of factual knowledge in their parameters~\citep{petroni2019lama,geva-etal-2021-transformer,roberts2020much}, which could potentially be out-dated or incorrect. Even for moderate-size models, it is prohibitively expensive to retrain every time an update happens or a mistake is uncovered in the model's parametric world knowledge. Even if resources are ample, it is non-trivial to ensure that the retraining only modifies the target without affecting other knowledge or skills present in the model. Furthermore, one piece of factual knowledge might have a large number of different mentions or it can be implicitly inferred from multiple sentences in the pretraining corpus, making it extremely difficult even to prepare an edited version of the training set.

\begin{table*}
    \scalebox{0.9}{
    \renewcommand{\arraystretch}{1.1}
    \centering 
    \begin{tabular}{@{}lp{7cm}p{7cm}@{}}
    \toprule
     & {\textbf{Controllability}} & {\textbf{Robustness}} \\ 
    \midrule
    \textbf{Question} & Dave Gilmour and Roger Waters were in which rock group? & How has \textbf{British art} survived in Normandy? \vspace{2mm} \\
    \midrule
    \textbf{Context} & George Roger Waters (born 6 September 1943) is an English singer, \dots Later that year, he reunited with \textbf{The Rolling Stones} bandmates Mason, Wright and David Gilmour for the Live 8 global awareness event; it was the group's first appearance with Waters since 1981… & In Britain, \textbf{Norman art} primarily survives as stonework or metalwork, such as capitals and baptismal fonts. In southern Italy, however, Norman artwork survives plentifully in forms strongly influenced by its Greek, Lombard, and Arab forebears. Of the royal regalia preserved in Palermo, the crown is Byzantine…\\ 
    \midrule
    \textbf{KAFT (ours)} & The Rolling Stones (from context). & In museums (irrelevant context). \vspace{2mm} \\
    \midrule
    \textbf{Noisy FT} & Pink Floyd & stonework or metalwork \vspace{2mm} \\
    \midrule
    \textbf{UQA V2 11B} & Pink Floyd & stonework or metalwork, such as capitals and baptismal fonts \vspace{2mm} \\
    \midrule
    \textbf{Pretrained} & Pink Floyd & As stonework and metalwork, such ascapi-tals and baptismal fonts \\
    \bottomrule
    \end{tabular}}
    \caption{Examples of model outputs demonstrating that, in contrast with baselines, a model obtained by KAFT is characterized by both 
    improved controllability by a context that contradicts its pretrained world knowledge, and improved robustness against an irrelevant context, compared to baseline methods. Here Pretrained refers to a T5 XXL model, which is also the underlying model for KAFT and Noisy Finetuning. UQA V2 11B is based on the T5 11B model.
    }
    \label{table:examples}
    \vspace{-3mm}
\end{table*}

In human cognition, working memory~\citep{Miller1960} provides the biological brain with the ability to hold information temporarily to perform tasks such as conversation, reasoning and mathematics in a way that is highly adaptive to the ever changing environment. As shown both experimentally and theoretically \citep{Fuster1973, Ashby2005}, working memory is stored in sustained activations of neurons, as opposed to the long term memory which is stored in weights. Working memory is also the immediate information buffer that is accessed while performing conscious tasks. In particular, it is where the fusion of perceptual inputs and long term memory happens~\citep{Fukuda2017}. This suggests that one potential method to solve LLMs' pointwise knowledge update and correction problem is to control the working memory stored in activations, rather than editing the long term memory stored in weights.

As demonstrated by their powerful in-context few shot learning abilities~\citep{gpt3}, LLM could utilize different activation patterns resulting from different contexts during inference to solve a diverse set of tasks without any changes in the weights. It is natural to expect that the same would be true with factual knowledge. In particular, one could prepare a large list of natural language statements covering desired knowledge updates and corrections. At inference time, one provides the relevant statements as context along with the input and hopes that the model would perform the task based on the new knowledge presented in this context.
Thus, if the model's working memory is indeed controllable by context, then a single model with static long term memory can produce different results based on a flexible set of factual knowledge available in different contexts. 
However, we demonstrate in this paper that this approach may fall short for many existing LLMs as they have greater tendencies to ignore the context and stick to their own pretrained world knowledge. This raises a natural question: 
\begin{center}
\textit{Is it possible to design a mechanism to ensure that the context can influence the model's working memory in a desirable manner?}
\end{center}

Note that any such mechanism has to take into account the possibility of encountering a noisy context. 
For example, any retrieval system that selects the task-relevant context from a large collection of contexts will be imperfect and occasionally provide irrelevant context.
In such cases, it's desirable that the model prediction does not get swayed by an irrelevant context. 
Interestingly, we show that the standard pretraining and finetuning methods do not ensure this behavior either. 
In fact, it's the noise encountered during the training that often leads to the model ignoring the context.

In this work, we provide an affirmative answer to the aforementioned question and propose a  novel approach -- \textit{knowledge-aware finetuning} (KAFT) -- to make an LLM's working memory truly controllable via \textit{relevant} context while ignoring the noisy or irrelevant context. 
Towards this, we aim to ensure that the model utilizes different types of information at its disposal in the following order:
\begin{align}
&\textit{relevant context}  \nonumber\\
& \quad {>} \textit{ model's pretrained world knowledge} \label{eq:priority1} \\
& \quad {>} \textit{ irrelevant context} \label{eq:priority2},
\end{align}
where $a>b$ indicates that $a$ is prioritized over $b$. Thus, if the model decides that the context is relevant, it should ground its output on the context, ensuring the \textit{controllability} of its working memory by context. This is crucial when the context is in conflict with the model's pretrained memory.
On the other hand, when the context is irrelevant, the model should instead stick to its pretrained world knowledge; thus ensuring \textit{robustness} of its working memory against noise.  \\

\noindent\textbf{Our contributions.}~We develop first LLMs that utilize different knowledge sources with a predefined order of priorities. Along the way, we develop a systematic understanding of the working memories of LLMs and identify their shortcomings. Our key contributions are summarized below.

\begin{enumerate}[leftmargin=0mm, listparindent=\parindent, itemsep=0mm, partopsep=0pt,parsep=0pt, itemindent=!]
    \item We undertake a systematic \textit{joint} study of both controllability and robustness of the working memory of LLMs. Focusing on question answering (QA) task, we define the context-question relevance based on whether the context entails an answer to the question. 
    We create a \textit{novel benchmark} to measure the controllability by including contexts that imply an answer which contradicts the model's pretrained knowledge.\footnote{We rely on in-context prompts in a {closed book} QA setup to measure the model's pretrained world knowledge. } Similarly, we propose a benchmark to measure robustness by introducing irrelevant contexts. We conduct an extensive evaluation of LLMs with different sizes across multiple architectures (encoder-decoder and decoder-only) and make the following key observations:
    \begin{enumerate}[leftmargin=5mm, itemsep=0mm, partopsep=0pt,parsep=0pt,itemindent=!,labelindent=3mm]
    \item \textit{LLMs could exhibit poor controllability.}~Our experiments consistently show that both pretrained and QA finetuned LLMs tend to igore a context when it contradicts with model's world knowledge. We show that this problem becomes more severe as the model becomes larger. We further show that the noise in the (QA) finetuning set plays an important role in emergence of this behavior. (Sec.~\ref{sec:Controllability})
    \item \textit{LLMs are not robust against context noise.}~We demonstrate that both pretrained and QA finetuned models are strongly interfered by irrelevant contexts, especially the ones that are on the same general topic as the underlying question. (Sec.~\ref{sec:Robustness})
    \end{enumerate}
    \item We propose a novel method -- knowledge aware finetuning (KAFT) -- to directly enhance both controllability (Eq.~\ref{eq:priority1}) and robustness (Eq.~\ref{eq:priority2}) of an LLM. KAFT enhances the controllability by creating counterfactual data augmentations where the answer entity in the context is swapped to a different but plausible entity, in conflict with the ground truth (and potentially the model's world knowledge). As for enhancing robustness, KAFT requires the model fit on to its pretrained closed-book answer rather than the ground truth answer whenever the context is irrelevant. 
    \item Through extensive empirical evaluation, we show that KAFT-based models successfully demonstrate the coexistence of controllability and robustness of model's working knowledge (see Table~\ref{table:examples} for an illustration).
\end{enumerate}

%% file: contributions.tex
%% file: related.tex
\section{Related Works}

\begin{table}[!t]
    \centering
    \scalebox{0.8}{
    \renewcommand{\arraystretch}{1.3}
    \begin{tabular}{@{}lcc@{}}
        \toprule
         & Robustness & Controllability \\
        \midrule
        Standard (noisy) finetuning & \xmark & \xmark \vspace{1mm} \\
        {\parbox{4.5cm}{Counterfactual~finetuning \\ \cite{longpre-etal-2021-entity}}} & \xmark & \cmark \vspace{1mm} \\
        KAFT (our work) & \cmark & \cmark \\
        \bottomrule
    \end{tabular}
    }
    
    \caption{Summary of our contributions.}
    \label{tbl:contributions}
\end{table}

\noindent \textbf{World knowledge in language models.}
Multiple recent works established that LLMs indeed utilize their parameters to memorize factual information present in their large pretraining corpus. In particular, \citet{petroni2019lama} utilize LAnguage Model Analysis (LAMA) probing to show that BERT models~\citep{devlin2018bert} act as a knowledge base by memorizing factual world knowledge. \citet{roberts2020much} establish the similar behavior for T5 models~\citep{raffel2019exploring}. Motivated by these, it is common practice to employ modern LLMs in tasks like closed book QA, which attest to the existence of the memorization of factual world knowledge by such models~\cite{palm}.\\ 

\noindent \textbf{Knowledge update in language models.} 
Given that most of the factual knowledge is ever-evolving, e.g., the current English Premier League winner can potentially change every year, the memorized outdated factual information or an unseen new fact may lead to an incorrect or poor prediction~\cite{lzaridou2021gap, onoe2022entity}. Furthermore, during model deployment, one may unearth certain undesirable outcomes and biases. As a naive strategy, one can frequently retrain a LM from scratch on the current snapshot of corpus (with outdated facts and problematic text removed) and ensure that model predictions are grounded in reality. However, this strategy is prohibitively expensive for LLMs; as a result, multiple recent efforts have focused on identifying how these models store the factual knowledge~\citep{geva-etal-2021-transformer} as well as devising efficient methods to update the specific knowledge stored in model parameters~\citep{zhu2020modifying,de2021editing,dhingra2022time,meng2022locating}. However, such strategies face the challenge that updating a particular factual knowledge may inadvertently affect other unrelated parametric knowledge. \citet{jang2022towards} propose a continual learning framework to update outdated knowledge and acquire new knowledge, while retaining the time-invariant knowledge. Furthermore, \citet{mitchell2022fast} present a method to edit models' prediction given an input-output pair. Unlike this line of work, we focus on updating the model behavior by providing a suitable context and ensuring that the model's working memory is controllable by such contexts. 

\noindent \textbf{Contextual and parametric knowledge.} Previous works utilized retrieved context for improving large language models to perform downstream tasks such as QA \cite{realm2020,joshi2020contextualized,petroni2020context}.
At the same time, LLMs memorize large amounts of knowledge in their parameters, most notably acquired during large scale pretraining. 
Despite this dichotomy, only a few studies addressed the relation between these two very different knowledge sources in the context of LLMs.
\citet{longpre-etal-2021-entity} finds that larger models have a greater tendency to ignore context in favor of the model's own parametric knowledge, and that the noise in the context in the finetuning set plays a big role in causing this behavior. 
We incorporate the algorithms proposed by \citet{longpre-etal-2021-entity} for mitigating this problem as baselines in Sec. \ref{sec:ablation} (the \emph{Noisy Finetuning} and \emph{Relevant Only Finetuning} approaches).
In a related work, \citet{kassner2019negated} showed that language models tend to be easily misled by certain types of irrelevant contexts. We observe similar phenomena in QA tasks and show that KAFT leads to more robust models against irrelevant contexts.
Finally, \citet{ContraQA2021} considers a very different relation between the model's world knowledge and the context, where the context may not be trustworthy and should be ignored by the model. 
Indeed, as one interesting extension for future work, one could consider to extend Eq.(\ref{eq:priority1}-\ref{eq:priority2}) to source1 > source2 > model's own knowledge > source3 > irrelevant contexts from all sources.

\emph{As we prepare the manuscript, we were made aware of an independent investigation by~\citet{disent-qa} that shares some important aspects of our work.}

%% file: methods.tex
\section{Methods}
\label{sec:methods}
 
For concreteness, let's consider a reading comprehension QA task where the model takes question $q$ together with a piece of context $c$ as its input. The question has an answer $a$. In addition, we also need a relevance label $r$ denoting whether the context entails the answer. 

Starting with a pretrained LM $M$, we would like to get a finetuned model $M^\prime$ such that when the context $c$ is relevant, its answer is always grounded on $c$, when $c$ is irrelevant, it sticks to the pretrained model's answer. In other words:
 
\begin{align}
r=1: & \hspace{1cm} M^\prime(c+q) = a  \\
r=0: & \hspace{1cm} M^\prime(c+q) = M(q) 
\label{eq:single_source_form}
\end{align}where $M$ is the pretrained model, $M^\prime$ is the finetuned model and $+$ denotes string concatenation. 

With this setup, we are establishing the priority order between knowledge sources, as per Eq.~(\ref{eq:priority1}-\ref{eq:priority2}). In particular, if there is a conflict between the relevant context and parametric knowledge, then the output should be consistent with the context. In addition, irrelevant context should have no influence on the model's output. Note that even though we are separating relevant vs irrelevant context here, the model does not know $r$ a priori. It has to determine $r$ based on the semantics of $c$ and $q$. 

For relevant or counterfactual context, the label is the ground truth or counterfactual answer, respectively. For empty or irrelevant context, the label is given by the pretrained model's answer to the same question in a few-shot closed book setting, reflecting the model's pretrained knowledge. To provide more interpretability, we make the model output its classification of the context's relevance along side the answer itself. See Table~\ref{table:kaft_output_format}.

\begin{table}
    \scalebox{0.9}{
    \renewcommand{\arraystretch}{1.1}
    \begin{tabular}{@{}lc@{}}
    \toprule
    {\textbf{Context type}} & {\textbf{Target sequence}} \\
    \midrule
    relevant context & {\parbox{4.5cm}{\centering\$\{ground truth answer\} \\ {(from context)}}} \vspace{2mm} \\ 
    \midrule
    irrelevant context & {\parbox{4.5cm}{\centering \$\{pretrained model's answer\} \\ {(irrelevant context)}}} \vspace{2mm} \\
    \midrule
    empty context & {\parbox{4.5cm}{\centering \$\{pretrained model's answer\} \\ {(empty context)}}} \vspace{2mm} \\
    \midrule
    counterfactual context & {\parbox{4.5cm}{\centering \$\{counterfactual answer\} \\ {(from context)}}} \\
    \bottomrule
    \end{tabular}}
    \caption{A summary of the output formats of the KAFT dataset.}
    \label{table:kaft_output_format}
    \vspace{-3mm}
\end{table}

\subsection{Datasets} 
We construct KAFT based on several public datasets, including SQuAD 2.0~\citep{squad2}, T-REx~\cite{trex}, QASC~\cite{qasc} and Trivia QA~\cite{triviaqa}. They cover several different QA formats, including multiple choice (QASC), Cloze (TReX), extractive (SQuAD) and open domain (TriviaQA).  For each dataset, we may construct different types of context and corresponding labels as summarized in Table~\ref{table:kaft_data}. 

\begin{table*}[h!]
    \scalebox{0.95}{
    \renewcommand{\arraystretch}{1.1}
    \centering 
    \begin{tabular}{@{}p{2cm}p{4.5cm}p{3.5cm}p{5cm}@{}}
    \toprule
    {\textbf{Dataset}} & {\textbf{Relevant Context}} & {\textbf{Irrelevant context}} & {\textbf{Counterfactual context}} \\ 
    \midrule
    TReX & Sampled irrelevant statements and one relevant statement & Sampled & Sampled irrelevant statements and one relevant statement with the answer entity replaced \vspace{2mm}\\  
    \midrule
    SQuAD 2.0 & From original dataset & Original human labeled and sampled & Relevant context with answer span replaced by counterfactual answer \vspace{2mm}\\  
    \midrule
    QASC & 2-stage retrieved statements and one golden statement & Sampled & None \vspace{2mm}\\  
    \midrule
   {Trivia-QA (wiki split)}
    & Retrieved contexts containing the answer and overlapping with the question & Retrieved contexts that do not contain the answer & Relevant context with answer span replaced by counterfactual answer \\  
    \bottomrule
    \end{tabular}}
    \caption{A summary of the construction of the KAFT data. For relevant context, the label is the ground truth answer; for counterfactual context, the label is the counterfactual answer; for irrelevant or empty context, the answer is the pretrained model's few shot closed book answer. All four datasets also include examples where no context is provided.}
    \label{table:kaft_data}
    \vspace{-3mm}
\end{table*}

\subsection{Models}
We select families of pretrained LLMs: T5 \citep{t5} representing the encoder-decoder architecture and PaLM \citep{palm} representing the decoder only architecture. We include all three PaLM models (8B, 62B and 540B) in our analysis, while with T5 we had to restrict to the largest sizes (XL and XXL, with 3B and 13B parameters respectively) because the smaller ones do not respond well to in-context few shot prompts, making it difficult to measure their pretrained world knowledge.

\subsection{Relevant context}
\label{sec:methods_relevant_context}
We define the relevance of a context by whether it logically entails an answer to the question. We emphasize the strong requirement of logical entailment here. In particular, even if a piece of context is on the same topic or the same entities as mentioned by the question, it might still be irrelevant by this definition. 
In practice, This happens often among retrieved results. 
In Sec~\ref{sec:ablation}, we show that if the model is still required to fit on to the ground truth label when given an irrelevant context, then the model becomes more likely to ignore relevant contexts. 

Therefore it is crucial to strive towards precise logical entailment when building relevant context. This is difficult with large scale datasets and even human raters make mistakes. We apply several techniques to improve the semantic connection between the context and the QA pair. 

SQuAD 2.0 has human labels for this particular aspect. But most datasets do not. For TReX, the question is cloze style where we mask a certain entity within the triplet statement. We build a relevant context by concatenating the original statements with a number of sampled irrelevant statements, after randomly shuffling the order of statements. This ensures the relevance of the context while keeping it challenging. The training set of QASC provides 2 gold statements that implies the answer via a two hop reasoning. We are using the 2-stage retrieved collection of statements similar to~\cite{uqa}. We find that the gold statements, or semantically equivalent ones, often exist in the retrieved results. To improve relevance we will randomly add one of the two golden statements and mix it in the retrieved context to build a relevant context for the KAFT training set.

Trivia QA is especially challenging because there is no human labeled gold context, while all existing contexts are obtained by a retrieval system. One might naively filter the context by whether they contain the answer. This turned out to be insufficient and leaves a large fraction of irrelevant contexts that do not logically entail the answer. We apply additional filters based on the unigram overlaps of the context with the question, as well as a filter on the output of a logically entailment model.

\subsection{Irrelevant Context}
An irrelevant context is any context that does not entail the answer. There is a difference of an "easy" vs "hard" irrelevant context. An easy irrelevant context is completely irrelevant, often discussing a different topic. A hard irrelevant context is on the same topic, sometimes discussing the same entities involved in the QA pair but does not logically entail the answer. 

It is easy to generate easy irrelevant contexts. We randomly sample other contexts in the same dataset to build irrelevant contexts for Trivia QA, QASC, TReX and (partly) SQuAD 2.0.

It is non-trivial to generate hard irrelevant contexts. SQuAD 2.0 already contains human labels on whether the answer can be derived from the context, thus providing hard irrelevant contexts. Trivia QA provides somewhat extensive paraphrases for each answer. Therefore we filter the retrieved contexts to find ones that does not contain any answer paraphrase, and use them as hard irrelevant context. 

\subsection{Probing pretrained knowledge with bulk inference}
\label{subsec:static}
We first use the pretrained model to generate $M(q)$ in Eq.~\ref{eq:single_source_form}, which will be used to assemble the KAFT finetuning dataset according to Eq.~\ref{eq:single_source_form}. We use hand-engineered few-shot knowledge probing prompts that condition the model to answer a question according to its world knowledge acquired during pretraining. In appendix.~\ref{sec:knowledge_probing_prompts}, we provide more details on the construction of these prompts.

\subsection{Counterfactuals}
\label{sec:counterfactuals}
To train the model to be controllable by the context, we explicitly engineer plausible training data where the context is in conflict with the model's pretrained world knowledge. Examples of such a datapoint can be found in Table~\ref{table:cf_gen}. 

\begin{table*}[h!]
    \scalebox{0.95}{
    \renewcommand{\arraystretch}{1.1}
    \centering 
    \begin{tabular}{@{}p{4.5cm}p{11.5cm}@{}}
    \toprule
    Question & In which country did Warsaw Pact officials meet to dissolve the alliance? \\ 
    \midrule
    Original answer & Hungary\\  
    \midrule
    Counterfactual answer & Russia    \\
    \midrule
    Original context & On 25 February 1991, the Warsaw Pact was declared disbanded at a meeting of defense and foreign ministers from remaining Pact countries
    meeting in \textbf{Hungary}.    \\
    \midrule
    Counterfactual context & On 25 February 1991, the Warsaw Pact was declared disbanded at a meeting of defense and foreign ministers from remaining Pact countries meeting in \textbf{Russia}.    \\
    \midrule
     T5 Prompt to generate the counterfactual answer & Let's play a game of writing fake answers Who did US fight in world war 1? Real answer: Germany. Fake answer: Somalia. Who is the
     CEO of Amazon? Real Answer: Jeff Bezos. Fake Answer: Richard D. Fairbank. \dots  \emph{7 more examples} \dots In which country did Warsaw Pact officials meet to dissolve
     the alliance? Real answer: Hungary. Fake answer: $\left< extra\_id\_0 \right>$. \\
    \bottomrule
    \end{tabular}}
    \caption{An example from the counterfactual split of the KAFT training set. We take an original question, answer and context triplet, using a few examples to prompt a pretrained T5 XXL model to generate a plausible counterfactual answer. We then replace all occurrences of the original answer with the counterfactual answer to build the counterfactual context.}
    \label{table:cf_gen}
    \vspace{-3mm}
\end{table*}

Given a triplet of question, answer and relevant context, we use a pretrained T5 XXL model to generate a triplet of question, counterfactual answer and counterfactual context. This is done in 3 steps: 1, we apply a diverse set of few-shot prompts similar to Table.~\ref{table:cf_gen} to condition a pretrained T5 XXL model to generate plausible counterfactual answers. 2, We remove examples if the generation is unsuccessful, when it's either too long or have a large overlap with the original answer. 3, We replace all occurrences of the original answer with the counterfactual answer in the original context to build the counterfactual context. With this approach, we build a new QA data set where the answer implied by the context is likely to be in conflict with the model's existing knowledge. 

\subsection{Metrics}
\label{sec:metrics}
In this section, we define metrics that measures controllability and robustness.
\paragraph{Controllability:} In control theory, controllability~\cite{controllability_book} refers to the ability of using external inputs to manipulate the system and reach all possible states. In the spirit of this definition, we measure the controllability of an LM's working memory by external contexts. We supply the model with a relevant counterfactual context and examine whether it can output the corresponding counterfactual answer. The counterfactual context is constructed using the method introduced in Sec.~\ref{sec:counterfactuals}. However we ensure no entities overlaps exist between the prompts that generates the training data vs the test data. For this work, we specifically select questions where all five pretrained models can answer correctly in a closed book few-shot setting, which are referred to as head questions. Since they are likely well represented in the pretraining set, such questions are the most challenging slice as we swap the answer to counterfactuals. Since we don't have any paraphrases of the counterfactual answer, we choose to use thresholded unigram recall to measure the performance. In particular, a model output is rated positive if the output of the model contains $>80\%$ of the answer unigrams, with stop-words removed.  

\paragraph{Robustness:} To measure robustness, we use the human labeled "impossible" slice of SQuAD 2.0, since SQuAD 2.0 contains many examples where the context is on the same general topic of the question but does not contain the answer. We measure the rate when the model successfully avoids extracting answers from such irrelevant contexts. The avoidance is considered successful if the context contains less than $50\%$ of the unigrams in the model's prediction, removing stop words. 

\subsection{Baselines}
\label{sec:baselines}
\paragraph{Pretrained:} We evaluate the pretrained model's ability to do zero shot reading comprehension QA with various types of contexts, which is concatenated with the question to build the input sequence to the model.
\paragraph{Noisy Finetuning:} The approach where the label is the ground truth answer whether the context is relevant or not. This is a very universal method and is the way most QA datasets are built.\footnote{As a notable exception, SQuAD 2.0 has empty strings as labels for its irrelevant context.} In this work, we construct this baseline for KAFT by first removing all counterfactual augmentations and then replace all labels with the ground truth label. 
\paragraph{Relevant Only Finetuning:} The approach where only relevant context and the corresponding ground truth label are used during finetuning, which is shown to improve controllability in \citep{longpre-etal-2021-entity}. As a baseline for KAFT we remove all counterfactual and irrelevant augmentations and only keep the relevant slice of our finetuning data. 
\paragraph{UQA V2:} The Unified QA 11B~\cite{uqav2} model, which is a general purpose QA model finetuned on a collection of 20 QA datasets. We take the largest model (11B) in the UQA V2 family as a baseline and compare with KAFT T5 XXL which is of similar size in~\ref{fig:data_eff}. Since UQA V2 contains SQuAD 2 in its training set, where the label for irrelevant context is the empty string, it is not completely noisy finetuned. 
\paragraph{KAFT noCF:} The KAFT method with no counterfactual augmentations. 
\paragraph{KAFT noCF and noTQA:} The KAFT method with no counterfactual augmentations and no trivia QA slice.

We include more details on the hyper parameters of model finetuning, prompts, post processing, data filtering and metric computations in the Appendix.

%% file: results.tex
\section{Results}
\subsection{Settings}
In this section we measure the controllability and robustness of KAFT with the metrics defined in Sec.~\ref{sec:metrics} and compare with baseline models and methods in Sec.~\ref{sec:baselines}. 
\begin{figure*}[h!]
    \centering
    \includegraphics[width=\textwidth]{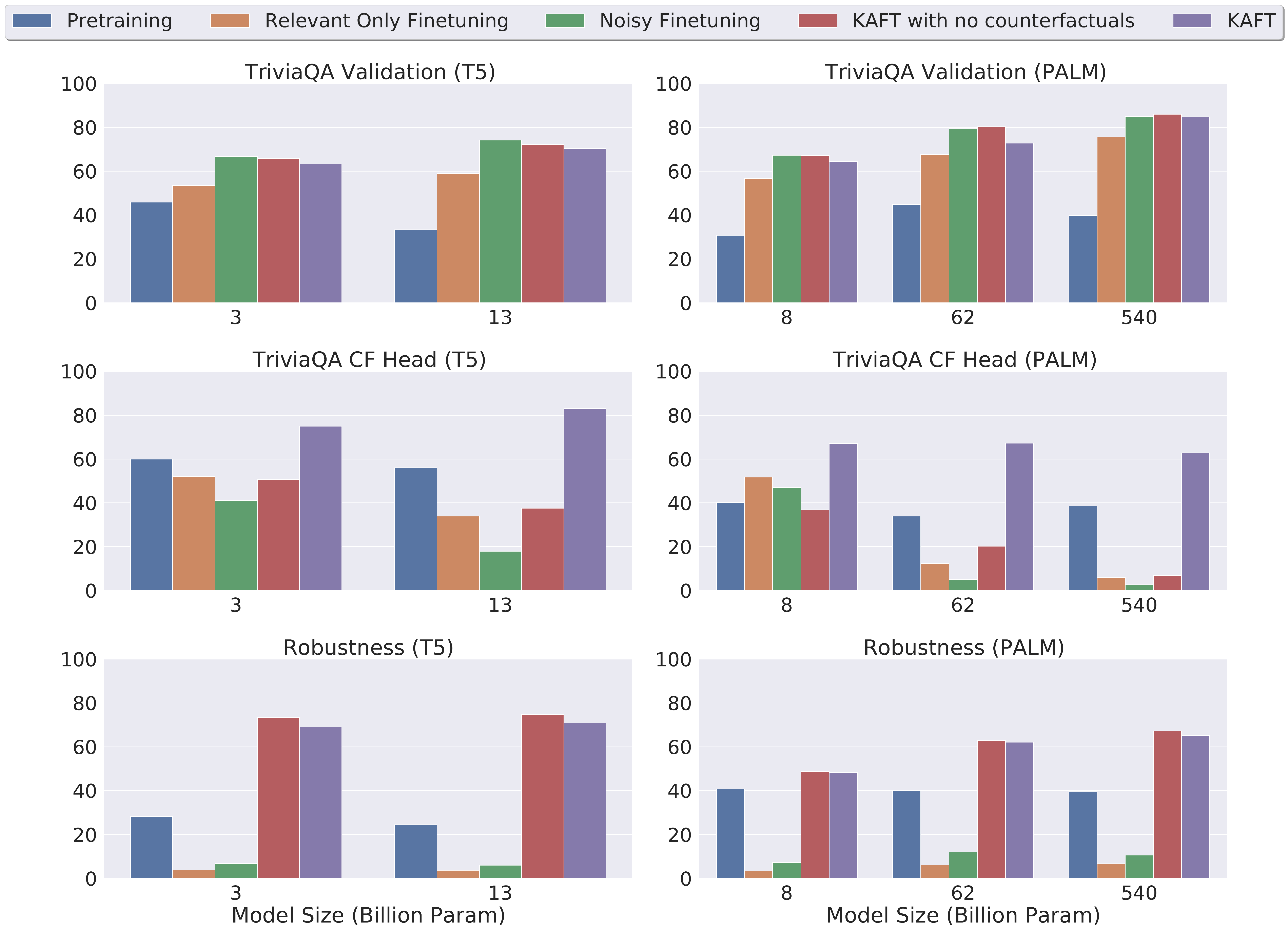}
    \caption{Large language models may become less controllable by context as the model size increases. Even when they obtain more world knowledge and become otherwise stronger. In the first row, we show the performance on the wiki split of Trivia QA when the model is provided one piece of context. On the second row, we show the model's controllability metric where a counterfactual trivia QA context is supplied. The third rows shows robustness metrics where a human labelled irrelevant context from SQuAD is supplied.}
    \label{fig:methods}
\end{figure*}

\subsection{Larger models are more likely to ignore contexts}
As a LM becomes larger, it becomes stronger at language understanding and obtains more entity-knowledge from pretraining. As a result, most benchmarks improve as a function of model size. In the first row of Fig.~\ref{fig:methods}, we demonstrate this effect on the validation set of Trivia QA, showing exact match accuracy. 

However, we found that larger models tends to ignore the context more. This happens both for the pretrained model as well as models finetuned on QA tasks using different approaches. We demonstrate this effect in the second row of Fig.~\ref{fig:methods} where the model is evaluated on contexts that contain counterfactual answer entities. Therefore, new methods are needed to improve the controllability of large language models. 

\subsection{KAFT and Controllability}
\label{sec:Controllability}
One of the most striking phenomenon observable from Fig.~\ref{fig:methods} is that KAFT achieve immense improvements on controllability while maintaining performance on regular datasets. For example, the KAFT PaLM 540B model achieves 24X better controllability compared to the noisy finetuning when the context is in conflict with the model's pretrained factual knowledge, while performing similarly on regular contexts. In addition, KAFT is the only finetuning approach that consistently achieves better controllability than the pretrained models. 

Perhaps not surprisingly, most of this gain originates from the counterfactual augmentation where the model explicitly learns the priority order Eq.~\ref{eq:priority1} when a conflict does appear. However it is worth noting that both relevant only finetuning and KAFT without counterfactual augmentations also exhibit stronger controllability compared to noisy finetuning, even when there is no explicit counterfactual augmentations in both cases. The reason is that both these approaches suppress the occurrence of cases where the context has no semantic link to an answer that was unknown to the model. Thus the model became less prone to ignore the context completely compared to noisy finetuning. 

\subsection{KAFT and Robustness}
\label{sec:Robustness}
Our observations on robustness are somewhat similar to controllability. One important difference is that there is no obvious improvement of robustness as model size increases: the robustness decreased slightly from T5 XL to XXL and from PaLM 8B to 62B. But the difference is small and there is no clear trend. Standard finetuning approaches severely reduce robustness. Relevant only finetuning suffers the most loss because it has not seen an irrelevant piece of context during training. Noisy finetuning only alleviates this loss slightly, still vastly under performing the pretrained model even when it has the same amount of irrelevant context in its training set compared to KAFT. 

KAFT, on the other hand, significantly boosts robustness. For example, the KAFT PaLM 540B model achieves 6X better robustness compared to noisy finetuning and 1.6X better robustness compared to the pretrained model. Adding the counterfactual augmentation slightly reduces robustness, but the difference is comparably small. 

\begin{figure}[h!]
    \centering
    \includegraphics[width=0.46\textwidth]{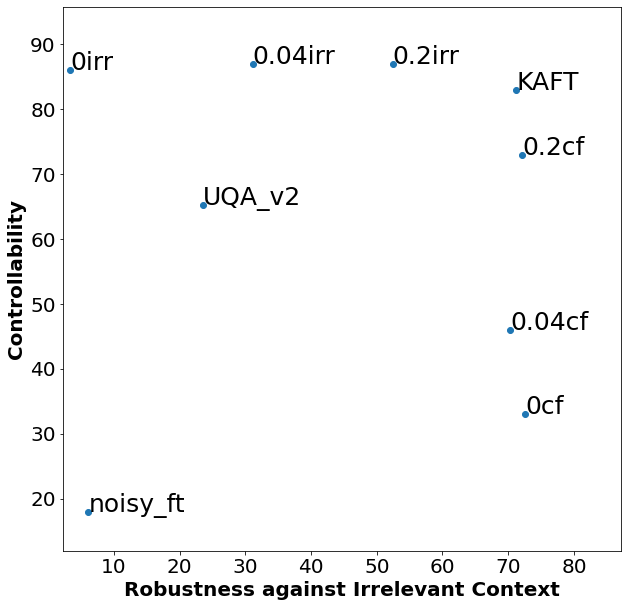}
    \caption{Ablation studies on data mixture ratios, showing the relative independence of controllability, robustness and standard metrics. Here e.g. 0.2irr refers to reducing the sampling rate of the irrelevant augmentation in KAFT to $20\%$; 0cf refers to removing all counterfactual augmentations from the KAFT datasets. We add UQA V2 and noisy finetuning baselines for comparison.} 
    \label{fig:data_eff}
\end{figure}

\subsection{Analysis and Ablation studies} 
\label{sec:ablation}
We perform two ablation studies to understand the effect of different augmentations in KAFT on controllability and robustness, as well as the general effect of added context noise. 

\noindent\textbf{Effect of KAFT data augmentations:}~In Fig.~\ref{fig:data_eff}, we systematically reduce the sampling rate of different data augmentation slices when training T5 XXL models. We observe that reducing or removing the counterfactual and irrelevant data augmentations severely impacts controllability and robustness, respectively. In addition, KAFT models significantly out-perform the very strong baselines of Unified QA V2 on both controllability and robustness, which is a general purpose QA model trained across a large collection of public QA datasets. Demonstrating that the KAFT method cannot be replaced by simply adding more supervised data.  \\

\noindent\textbf{KAFT models do not memorize counterfactual:}~One potential danger of adding counterfactual context-answer pairs in the training set is unwanted memorization. We check whether the KAFT model memorizes the counterfactual answers in the training set using the same prompts we used to probe the pretrained model's closed book answers. The results in Table.~\ref{table:cf_memorization} shows that KAFT has little unwanted memorization of counterfactual answers. Instead the model learns the desirable correlation between the context and the output, as demonstrated in Fig~\ref{fig:methods}. \\

\noindent\textbf{Context noise reduces controllability:}~Here by context noise we refer specifically to the subset of training data where the model is required to produce an answer that is not implied by the provided context, or required to ignore the context while it actually imply the answer. On the flip side, we find that it is possible to achieve good controllability without explicit counterfactual augmentations if we can reduce context noise in the training data.

In particular, because trivia QA contexts are produced by a retrieval system, it is not guaranteed that a context logically implies the answer. This is even true when the context contains exact matches of the answer. On the other hand, TReX, SQuAD and QASC contains much less context noise given the our KAFT construction methods Sec.~\ref{sec:methods_relevant_context}. Due to this intrinsic noise, including trivia QA in KAFT caused a negative impact on controllability, especially when there are no explicit counterfactual augmentations. Table.~\ref{table:noise_eff_CF} shows how different amounts of context noise impact the model's controllability. The first row shows noisy finetuning, which contains the most noise. The last row shows that KAFT with Trivia QA data removed. Even though this model is not finetuned on Trivia QA, it has the best controllability compared to other methods. The second row uses a simpler and more noisy filter than Sec.~\ref{sec:methods_relevant_context} by considering a context to be relevant if it contains exact matches to the answer.

\begin{table}[h!]
    \begin{center}
    \scalebox{1.0}{
    \renewcommand{\arraystretch}{1.15}
    \begin{tabular}{@{}lcc@{}}
    \toprule
    \multirow{1}{*}{\textbf{Model}} & {\textbf{Pretrained}} & {\textbf{KAFT}} \\
    \midrule
    T5 XL & $6.1\%$ & $7.2\%$ \\
    T5 XXL & $6.6\%$ & $6.8\%$ \\
    PaLM 8B & $3.3\%$ & $4.1\%$ \\
    PaLM 62B & $1.4\%$ & $1.3\%$ \\
    PaLM 540B & $0.6\%$ & $0.7\%$ \\
    \bottomrule
    \end{tabular}
    }
    \end{center}
    \caption{The match rate between models' closed book answers and counterfactual answers, among all TriviaQA training set questions with counterfactual augmentations. KAFT shows little unwanted memorization of counterfactual answers.}
    \label{table:cf_memorization} 
    \vspace{-3mm}
\end{table}

\begin{table}
    \scalebox{0.92}{
    \renewcommand{\arraystretch}{1.15}
    \begin{tabular}{@{}lcc@{}}
    \toprule
    \multirow{2}{*}{\textbf{Method}} & {\textbf{TQA-CF}} & {\textbf{TQA-CF}} \\
     & \textbf{PALM 62B} & \textbf{T5 XXL} \\
    \midrule
    NoisyFT & $15\%$ & $37\%$ \\
    KAFT noCF EM filter & $20\%$ & $51\%$ \\
    KAFT noCF & $33\%$ & $54\%$ \\
    KAFT noCF and noTQA & $52\%$ & $69\%$ \\
    \bottomrule
    \end{tabular}}
    \caption{We compare the controllability on the head counterfactual questions for finetuning methods with different levels of context noise, which increases from the first row to the last. Context noise leads to model ignoring context and thus lower controllability.}
    \label{table:noise_eff_CF} 
    \vspace{-3mm}
\end{table}

%% file: conclusion.tex
\section{Conclusion}
In this work, we analyzed the interaction between the pretrained world knowledge of LLMs and knowledge contained in informational contexts provided as a part of the input sequence. We find that models are prone to ignore the context, especially when the context are in conflict with the model's internal knowledge. In addition, we find that the model's output can be swayed by irrelevant context even when there is no logical link between such context and the model's task at hand. We characterize these as controllability and robustness issues of large language models when one attempts to control its working memory with noisy context. We proposed a new finetuning method, KAFT, that contains various data augmentations that substantially boost the controllability and robustness of a LLM while does not significantly affect its performance on regular metrics. With KAFT, we can build LLMs with a clear order of priority when utilizing information from difference sources, including its own pretrained world knowledge. 

%% file: future_work.tex
\section{Future work}

\subsection{Multiple Sources}
In this work, we trained a model that can utilize two sources of information with predefined priority order, with one of them being the model's own parametric knowledge. In future, this can be expanded to multiple sources of different quality or trustworthiness: 
\begin{align}
& \text{relevant context 1} > \text{relevant context 2} \\ 
& > \text{model's parametric knowledge} \\
& > \text{relevant context 3} \\
& > \text{all irrelevant context}
\end{align}This orders of priority determines the handling of conflicts. In addition, any irrelevant context should have no influence on the model's output.

\subsection{Dynamically enforce "learning to ignore"}
\label{subsec:dynamic}
In this work, it was necessary to build a different KAFT dataset for each model. Because in Eq.~\ref{eq:single_source_form}, whenever the context is irrelevant, the model fits on to the pretrained model's answers which is different for each model.
In future we'd like to explore a dynamic methods that generates closed booked answers during training.
To do this, at each training step involving irrelevant context, we will run the forward pass twice, one with the provided context and another without. Then we can compute a new loss: 
\begin{align}
r=1: &\hspace{0.1cm} \text{Loss} = \text{CE}(M^\prime(c+q), \text{label})  \\
r=0: & \hspace{0.1cm} \text{Loss} = \text{CE}(M^\prime(c+q), \\
        & \hspace{1.5cm} \text{stop\_gradient}(M^\prime(q))) 
\label{eq:dynamic}
\end{align} where $+$ denotes string concatenation. This is different from Eq.~\ref{eq:single_source_form} as it fits on to the closed book answers of the current version of the finetuned model, rather than that of the pretrained model. It is not yet clear whether this approach would achieve better robustness. It is also more expensive because two forward passes are necessary for each training example. However it might be justified by the improved simplicity in directly applying KAFT on a dataset with ground truth labels and context relevance labels with minimal prepossessing. 

This approach is somewhat similar to classifier free guidance~\cite{classifier_free_guidance}, which has been successfully applied to image generation models. One added benefit of classifier free guidance is the ability to tune the strength of context conditioning after the model is trained, which is another interesting direction to explore here.  

%% file: appendix.tex
\section{Appendix}
\label{sec:appendix}
\subsection{Training Details}
We use a learning rate of 0.0002 on all models. The batch size is 32 for all PaLM models and 16 for T5 models. For T5 XL we pick the checkpoint at 100000 finetune steps and for T5 XXL models we pick the checkpoint at 90000 steps. For PaLM 8B and 62B, we pick the checkpoint at 40000 finetuning steps. For PaLM 540B we pick the checkpoint at 15000 steps. These steps are generally determined by avoiding overfitting. However for larger models we are also contrained by compute resources.

\subsection{Knowledge Probing Prompts}
\label{sec:knowledge_probing_prompts}
In this section we provide details on how the knowledge probing prompts in Table.~\ref{table:knowledge_probing_prompts_squad}-\ref{table:knowledge_probing_prompts_qasc} are constructed. In particular, our goal is to make the model only answer questions where it knows the answer. To do this, we construct prompts that contains two types of QA pairs: 1) Regular QA pairs if the model can answer the specific question correctly in multiple few-shot in-context settings. 2) QA pairs where the answer is "I don't know" for T5 models or "?" for PaLM models, if the model cannot answer the question correctly in most few-shot in-context settings. With such specially designed prompts, we encourage the model to abstain if it does not know the answer. 

\begin{table*}
    \centering 
    \scalebox{0.7}{
    \renewcommand{\arraystretch}{1.1}
    \begin{tabular}{@{}ll@{}}
    \toprule
    Model & Standard QA Knowledge Probe Prompts \\ 
    \midrule
    T5 XL & \makecell[l]{Q: Into what body of water does the Hudson River terminate? A: The Atlantic Ocean.\\Q: What method formally adds inverses to elements to any monoid? A: I don't know.\\Q: Supply and what else causes child labour to still exist today? A: demands.\\Q: Who is the prime minister of Japan in 2015? A: Shinzo Abe.\\Q: Who is responsible for judicial review? A: Courts.\\Q: what was the name of the other HD channel Virgin media could carry in the future? A: I don't know.\\Q: What is the term for a hyperactive immune system that attacks normal tissues? A: autoimmunity.\\Q: What complexity class is commonly characterized by unknown algorithms to enhance solvability? A: I don't know.\\Q: Which nation contains the majority of the amazon forest? A: Brazil.}  \\
    \midrule
    T5 XXL & \makecell[l]{Q: Into what body of water does the Hudson River terminate? A: The Atlantic Ocean.\\Q: What method formally adds inverses to elements to any monoid? A: I don't know.\\Q: Supply and what else causes child labour to still exist today? A: demands.\\Q: Who is the prime minister of Japan in 2015? A: Shinzo Abe.\\Q: Who is responsible for judicial review? A: Courts.\\Q: What religion did the French spread along with their imperialism?  A: Catholicism.\\Q: The symbol for mercuric oxide is? A: HgO.\\Q: What religion did the Yuan discourage, to support Buddhism? A: Taoism.}  \\
    \midrule
    PaLM 8B & \makecell[l]{Only answer the questions you know the answer to:\\Q: Into what body of water does the Hudson River terminate? A: The Atlantic Ocean.\\Q: What year was the county of Hampshire officially named? A: ?.\\Q: Who said the following statement? "Enlightenment is man\'s emergence from his self-incurred immaturity". A: Immanuel Kant.\\Q: What method formally adds inverses to elements to any monoid? A: ?.\\Q: What King and former Huguenot looked out for the welfare of the group? A: Henry IV.\\Q: The principle of faunal succession was developed 100 years before whose theory of evolution? A: Charles Darwin.\\Q: Who is the hero who killed a dragon on the Drachenfels? A: Siegfried.} \\
    \midrule
    PaLM 62B & \makecell[l]{Only answer the questions you know the answer to:\\Q: Into what body of water does the Hudson River terminate? A: The Atlantic Ocean.\\Q: What year was the county of Hampshire officially named? A: ?.\\Q: Who said the following statement? "Enlightenment is man's emergence from his self-incurred immaturity". A: Immanuel Kant.\\Q: What method formally adds inverses to elements to any monoid? A: ?.\\Q: Who was the US Secretary of State in 2001? A: Colin Bowell.\\Q: The principle of faunal succession was developed 100 years before whose theory of evolution? A: Charles Darwin.\\Q: Who is the hero who killed a dragon on the Drachenfels? A: Siegfried.\\Q: When did the European Anti-Fraud Office investigate John Dalli? A: 2012.\\Q: What religion did the French spread along with their imperialism?  A: Catholicism.\\Q: When did Costa v ENEL take place? A: 1964.} \\ 
    \midrule
    PaLM 62B & \makecell[l]{Only answer the questions you know the answer to:\\Q: Into what body of water does the Hudson River terminate? A: New York Bay.\\Q: What year was the county of Hampshire officially named? A: ?.\\Q: Who said the following statement? "Enlightenment is man\'s emergence from his self-incurred immaturity". A: Immanuel Kant.\\Q: What method formally adds inverses to elements to any monoid? A: ?.\\Q: When was the Parental Leave directive created? A: 1996.\\Q: How many megaregions are there in the United States? A: 11.\\Q: Where is D\'Olier Street? A: Dublin.\\Q: What is the speed limit set to reduce consumption? A: 55 mph.\\Q: What channel replaced Sky Travel? A: Sky Three.\\Q: Who founded McKinsey \& Company? A: James O. McKinsey.} \\
    \bottomrule
    \end{tabular}}
    \caption{Knowledge probing prompts for standard QA datasets. These prompts are used to probe the pretrained model's answer to questions in SQuAD 2.0 and Trivia QA.}
    \label{table:knowledge_probing_prompts_squad}
    \vspace{-3mm}
\end{table*}

\begin{table*}
    \scalebox{0.7}{
    \renewcommand{\arraystretch}{1.1}
    \centering 
    \begin{tabular}{@{}l@{\qquad}l@{}}
    \toprule
    Model & Cloze Style QA Knowledge Probe Prompts \\ 
    \midrule
    T5 XL & \makecell[l]{The Hudson River terminate into \_\_\_ . A: The Atlantic Ocean.\\\_\_\_ formally adds inverses to elements to any monoid. A: ?.\\Supply and \_\_\_ causes child labour to still exist today? A: demands.\\\_\_\_ was the prime minister of Japan in 2015? A: Shinzo Abe.\\\_\_\_ is responsible for judicial review. A: Courts.\\\_\_\_ was the name of the other HD channel Virgin media could carry in the future. A: ?.\\\_\_\_ is defined as a hyperactive immune system attacking normal tissues? A: autoimmunity.\\\_\_\_ complexity class is commonly characterized by unknown algorithms to enhance solvability. A: ?.\\\_\_\_ contains the majority of the amazon forest? A: Brazil.} \\
    \midrule
    T5 XXL & \makecell[l]{The Hudson River terminate into \_\_\_ . A: The Atlantic Ocean.\\\_\_\_ formally adds inverses to elements to any monoid. A: ?.\\Supply and \_\_\_ causes child labour to still exist today? A: demands.\\\_\_\_ was the prime minister of Japan in 2015? A: Shinzo Abe.\\\_\_\_ is responsible for judicial review. A: Courts.\\The French spread along with their imperialism the \_\_\_ religion. A: Catholicism.\\The symbol for mercuric oxide is \_\_\_. A: HgO.\\The Yuan discouraged \_\_\_ to support Buddhism. A: Taoism.}  \\
    \midrule
    PaLM 8B & \makecell[l]{Only answer the questions you know the answer to:\\The Hudson River terminate into \_\_\_ . A: The Atlantic Ocean.\\The county of Hampshire was officially named in \_\_\_ . A: ?.\\\_\_\_ said "Enlightenment is man\'s emergence from his self-incurred immaturity". A: Immanuel Kant.\\\_\_\_ formally adds inverses to elements to any monoid. A: ?.\\King \_\_\_ and former Huguenot looked out for the welfare of the group. A: Henry IV.\\The principle of faunal succession was developed 100 years before \_\_\_'s theory of evolution. A: Charles Darwin.\\\_\_\_ is the hero who killed a dragon on the Drachenfels? A: Siegfried.} \\
    \midrule
    PaLM 62B & \makecell[l]{Only answer the questions you know the answer to:\\The Hudson River terminate into \_\_\_ . A: The Atlantic Ocean.\\The county of Hampshire was officially named in \_\_\_ . A: ?.\\\_\_\_ said "Enlightenment is man\'s emergence from his self-incurred immaturity". A: Immanuel Kant.\\\_\_\_ formally adds inverses to elements to any monoid. A: ?.\\\_\_\_ was the US Secretary of State in 2001. A: Colin Bowell.\\The principle of faunal succession was developed 100 years before \_\_\_'s theory of evolution? A: Charles Darwin.\\\_\_\_ is the hero who killed a dragon on the Drachenfels. A: Siegfried.\\The European Anti-Fraud Office investigate John Dalli in year \_\_\_ . A: 2012.\\The French spread along with their imperialism the \_\_\_ religion. A: Catholicism.\\Costa v ENEL happend in year \_\_\_ . A: 1964.} \\ 
    \midrule
    PaLM 62B & \makecell[l]{Only answer the questions you know the answer to:\\The Hudson River terminate into \_\_\_ . A: New York Bay.\\The county of Hampshire was officially named in \_\_\_ . A: ?.\\\_\_\_ said "Enlightenment is man\'s emergence from his self-incurred immaturity". A: Immanuel Kant.\\\_\_\_ formally adds inverses to elements to any monoid. A: ?.\\The Parental Leave directive created in year \_\_\_ . A: 1996.\\ There are \_\_\_ megaregions in the United States. A: 11.\\D'Olier Street is located in \_\_\_ . A: Dublin.\\The speed limit was set to \_\_\_ to reduce consumption. A: 55 mph.\\\_\_\_ channel replaced Sky Travel. A: Sky Three.\\\_\_\_ founded McKinsey \& Company. A: James O. McKinsey.} \\
    \bottomrule
    \end{tabular}}
    \caption{Knowledge probing prompts for Cloze style QA datasets. These prompts are used to probe the pretrained model's answer to questions in TReX.}
    \label{table:knowledge_probing_prompts_trex}
    \vspace{-3mm}
\end{table*}

\begin{table*}
    \scalebox{0.8}{
    \renewcommand{\arraystretch}{1.1}
    \centering 
    \begin{tabular}{@{}lc@{}}
    \toprule
    Model & Multiple Choice QA Knowledge Probe Prompts \\ 
    \midrule
    PaLM 62B & \makecell[l]{Question: Into what body of water does the Hudson River terminate? (A) The great lakes \\ 
    (B) Amazon river (C) The red sea (D) the Atlantic Ocean (E) San Francisco bay \\ 
    (F) The north sea (G) Indian Ocean (H) Lake Mississippi -Answer: (D) the Atlantc Ocean. \\
    Question: Who was the prime minister of Japan in 2015? (A) Donald Trump (B) Miho Nonaka \\
    (C) Andrew Yang (D) a France citizen (E) a political outsider (F) Shinzo Abe (G) woman \\
    (H) Zoe. -Answer: (F) Shinzo Abe.Question: what increases moisture? (A) density (B) the sun\\
    (C) wind (D) droughts (E) Honey (F) 17 (G) rain (H) meat -Answer: (G) rain. \\ 
    Question: What can be found inside a cell? (A) soil (B) dogs (C) ovum (D) starfish\\
    (E) Most plants (F) RNA (G) washer (H) abundant -Answer: (F) RNA.\\
    Question:What kind of coloring do chomoplasts make? (A) fat (B) move\\
    (C) RNA (D) grow (E) red (F) skin (G) eyes (H) DNA -Answer: (E) red.} \\
    \bottomrule
    \end{tabular}}
    \caption{Knowledge probing prompts for Cloze style QA datasets. These prompts are used to probe the pretrained model's answer to questions in TReX.}
    \label{table:knowledge_probing_prompts_qasc}
    \vspace{-3mm}
\end{table*}

\subsection{Postprocessing}
\label{subsec:postprocessing}
After we obtain the output from the pretrained model to the question, which is concatenated after the knowledge probing prompt, we need to postprocess it and removed unwanted components. We do two types of post-processing on the pretrained predictions: 
\begin{enumerate}
\item \textbf{Truncation:} We truncate the model's output on special tokens such as $<extra\_id\_1>$, punctuation, line change symbols and question/context initialization symbols such as "Q:", "Question:", "CONTEXT:". These symbols are a frequent in the pretrained model's responds to our QA style knowledge probe prompts and indicate that the model is ready to move on to the next question that is unrelated to the answer of the current question.  
\item \textbf{Abstain:} We normalize all abstain symbols. Whenever the model indicate abstaining using either "I don't know", "unsure" or "?" in the output as responses to our prompt, we record "unsure" as its answer when constructing the label in the irrelevant slices of KAFT.
\end{enumerate}

\subsection{Mixture weights}
KAFT mixes together a number of datasets, each with multiple augmentation slices. During training, data from these difference sources are sampled round-robin style according to predefined mixture weights. We list these weights as well as the corresponding dataset stats as in Table.\ref{table:task_weights}. The sampling ratio from each slice is computed using a product of the normalized dataset level rate and the normalized slice level rate as follows: 
\begin{equation}
    R(d, s) = \frac{r_d}{\sum_{d^\prime} r_{d^\prime}} \frac{r_{ds}}{\sum_{s^\prime} r_{ds^\prime}}
\end{equation}
where $d, d^{\prime}$ denote different datasets and $s, s^\prime$ denote difference slices within each dataset. For example, the sampling ratio from the QASC relevant slice is given by:
\begin{align}
    &R(QASC, relevant) \\
    &= \frac{0.3}{1.3+0.3+0.1+0.2} \frac{0.5}{0.5+0.25+0.02} \\
    &= 0.0831
    \label{eq:data_rates}
\end{align}

\begin{table*}
    \centering
    \scalebox{0.8}{
    \renewcommand{\arraystretch}{1.1}
    \begin{tabular}{@{}p{2cm}p{2cm}p{5cm}p{2cm}@{}}
    \toprule
    {\textbf{dataset}} & {\textbf{dataset weight}} & {\textbf{slice}} & {\textbf{slice weight}}\\
    \midrule
    SQuAD 2.0 & 1.3 & relevant & 0.8 \\
     & & counterfactual & 0.1 \\
     & & original irrelevant abstain & 0.1 \\
     & & original irrelevant other & 0.1 \\
     & & empty correct & 0.33 \\
     & & empty abstain & 0.02 \\
     & & empty other & 0.05 \\
     & & sampled irrelevant correct & 0.33 \\
     & & sampled irrelevant abstain & 0.02 \\
     & & sampled irrelevant other & 0.03 \\
    \midrule
    QASC & 0.3 & relevant & 0.5 \\
    & & irrelevant correct & 0.25 \\
    & & irrelevant other & 0.02 \\
    \midrule
    TReX & 0.1 & relevant & 0.4 \\
    & & counterfactual & 0.4 \\
    & & 2-hop relevant & 6 \\
    & & irrelevant correct & 0.15 \\
    & & irrelevant abstain & 0.03 \\
    & & irrelevant other & 0.03 \\
    \midrule
    Trivia QA & 0.2 & relevant & 0.8 \\
    & & counterfactual & 0.15 \\
    & & irrelevant/empty correct & 0.5 \\
    & & irrelevant/empty other & 0.2 \\
    \bottomrule
    \end{tabular}}
    \caption{Task mixture weights. During finetuning, training data from each split is computed round robin according to these weights. The sampling rate from each slice is comuted with these weights using in Eq.\ref{eq:data_rates}. Here "relevant", "irrelevant", "empty" indicates the relevance (or absence) of the context relative to the question. "counterfactual" indicates counterfactual context constructed using answer replacement. The additional specification for irrelevant/emtpy slices, "correct", "abstain" and "other" indicate the pretrained model's answers' type and quality relative to the ground truth. For TReX, we have a special slice called "2-hop relevant". These are relevant contexts contructed using 2-hop reasoning over the triplet structure of TReX.}
    \label{table:task_weights}
    \vspace{-3mm}
\end{table*}